\documentclass{sig-alternate}

\usepackage{graphicx}
\usepackage{amsfonts}
\usepackage{amsmath}
\usepackage{color}
\usepackage{url}
\usepackage{booktabs}
\usepackage{multicol}
\usepackage{times}
\usepackage[ruled,vlined]{algorithm2e}
\usepackage{dcolumn}
\usepackage{cleveref}

\usepackage{balance}

\DeclareGraphicsExtensions{.eps}

\begin{document}


\title{Local Optima Networks and \\ the Performance of  Iterated Local Search}

\numberofauthors{4} %
\author{
%
%
\alignauthor
Fabio Daolio\\
       \affaddr{Faculty of Business and Economics, DESI}\\
       \affaddr{University of Lausanne}\\
       \email{fabio.daolio@unil.ch}
\alignauthor
S\'ebastien Verel\\
       \affaddr{Université Nice-Sophia Antipolis}\\
       \affaddr{INRIA Lille Nord Europe}\\
       \email{verel@i3s.unice.fr}
\and  
\alignauthor
Gabriela Ochoa\\
       \affaddr{School of Computer Science, A.S.A.P. Group}\\
       \affaddr{University of Nottingham}\\
       \email{gxo@cs.nott.ac.uk}
\alignauthor
Marco Tomassini\\
       \affaddr{Faculty of Business and Economics, DESI}\\
       \affaddr{University of Lausanne}\\
       \email{marco.tomassini@unil.ch}
}

\maketitle

\begin{abstract}
Local Optima Networks (LONs) have been recently proposed as an alternative model of combinatorial fitness landscapes. The model compresses the information given by the whole search space into a smaller mathematical object that is the graph having as vertices the local optima and as edges the possible weighted transitions between them. A new set of metrics can be derived from this model that capture the distribution and  connectivity of the local optima in the underlying configuration space.This paper  departs from the descriptive analysis of  local optima networks, and actively studies the correlation between network features and the performance of a  local search heuristic. The $NK$ family of landscapes and the Iterated Local Search metaheuristic are considered.   With a statistically-sound approach based on multiple linear regression, it is shown that some  LONs' features strongly influence  and can even partly predict the performance of  a heuristic search algorithm. This study validates the expressive power of  LONs   as a model of combinatorial fitness landscapes.

\end{abstract}




\section{Introduction}
\label{sec:intro}

One of the most conspicuous limitations of heuristic search methods in combinatorial optimization, is the ability to become trapped at a local optimum~\cite{glover1986future}. The number and distribution of local optima in a search space have, therefore, an important impact on the performance of heuristic search algorithms. A recently proposed model: Local Optima Networks~\cite{pre08,verel2010local}, provides an intermediate level of description for combinatorial fitness landscapes. The model has a higher descriptive power than a single statistical metric or set of metrics; but it also compresses the search space into a more manageable mathematical object. Specifically, a graph having as vertices the optima configurations of the problem and as edges the possible weighted transitions between these optima.  This network representation allows the application of new analytical tools and metrics to study combinatorial landscapes, namely, those of complex networks analysis (e.g. degree distribution, clustering coefficient, assortativity, and community structure, to name a few).  In previous work, alternative definitions of edges have been studied, and some of these metrics  have been computed on the network extracted for two combinatorial problems:  the NK family of landscapes~\cite{pre08,gecco08,ppsn10,verel2010local,verellocal}, and the Quadratic Assignment problem~\cite{daolio2010local,daolio2011communities}.  Those studies have been mainly descriptive, although distinctive correlations between some network features and previous knowledge about search difficulty in these landscapes had been found. A previous related work  \cite{Pelikan:2010:NLP:1830483.1830606}, explored the relationships between $NK$ landscape features and the performance of a hybrid EA. The authors use standard landscape metrics, and conduct a study based mainly on scatter plots. They suggest that: ``further work is necessary to gain better understanding of the escape rate on the actual problem difficulty''. The present study, addresses exactly this point. The goal is to systematically explore correlations between local optima network features and the performance of a stochastic local search algorithm (Iterated Local Search) running on the underlying combinatorial optimization problem (in this study the $NK$ family of landscapes). The `escape rate' is a property related to the local optimum, and the LON could be considered as a new tool to better understand problem difficulty. The ultimate goal is to have predictive models of the performance of specific local search heuristics when solving a given combinatorial optimization problem.  This paper proposes an initial predictive model of performance based on the most influential LON features.

\section{Methods}
\label{sec:methods}
The local optima network model of combinatorial landscapes, and the iterated local search metaheuristic are considered in this study. The relevant definitions and experimental setup are given below.

\subsection{Local Optima Network}
\label{sec:LON}

A \emph{fitness landscape}~\cite{reidys2002combinatorial} is a triplet $(S, V, f)$ where $S$ is a set of potential solutions i.e. the \emph{search space}, $V : S \longrightarrow 2^S$ is a function that assigns to every $s \in S$ a set of neighbors $V(s)$ i.e the \emph{neighborhood structure}, and $f : S \longrightarrow \mathbb{R}$ is the evaluation of the corresponding solutions i.e. the \emph{fitness function}.

The present study uses the well-known $NK$-landscapes~\cite{kauffman93} as a benchmark set.
It is a problem-independent model for constructing combinatorial landscapes that are tunably rugged. In the model, $N$ refers to the number of (binary) genes in the genotype, i.e. the string length, and $K$ to their epistatic interaction, i.e. the number of other loci (chosen at random here) that influence the fitness contribution of a particular gene. Starting from this $N$-loci $2$-allele additive model, by increasing the non-linearity $K$ from 0 to $N-1$, the landscapes can be tuned from smooth to rugged.
Hence, $S$ is the search space of all $N$-bit binary strings, and its size is $\sharp S=2^N$. The neighborhood is defined by the minimum possible move on it, which is the single bit-flip operation, and the neighborhood size is $\sharp V(s) = N$. The fitness function evaluates each genotype $s$ as
$f(s)=\frac{1}{N}\sum_{i=1}^{N}f_i(s_i,s_{i1},\ldots,s_{iK})$,
where the values of loci contributions $f_i: \{0,1\}^{K+1}\rightarrow [0, 1]$ are drawn uniformly at random in $[0,1]$.

A \emph{local optimum} ($LO$), which is taken to be a maximum here, is a solution $s^{*}$ such that $\forall s \in V(s)$, $f(s) \leq f(s^{*})$. All optima are determined through exhaustive search by recursively running the 1-bit-flip best-improvement hill-climber, as in Algorithm~\ref{alg:HC}.
Let us denote by $h(s)$ the operator that associates to each solution $s \in S$, the solution obtained after applying that algorithm until convergence to a $LO$. Since $S$ is of finite size and there is no neutrality in $f(s)$ values, this produces a partition of the landscape in a finite number of basins of attraction, which we can denote by $LO_1$, $LO_2$, $LO_3 \ldots, LO_{nv}$, the local optima.

\begin{algorithm}[!h]
\caption{Best-Improvement Hill-Climber} \label{alg:HC}
Choose initial solution $s \in S$ \;
\Repeat{$s$ is a  Local Optimum}{
	choose $s' \in V(s)$, such that $f(s') = max_{x \in V(s)} f(x)$\;
	\If{$f(s) < f(s')$}{
		$s \leftarrow s'$\;
       }
}
\end{algorithm}

The connections among them account for the chances of escaping from a $LO$ and jumping into another one with a controlled move~\cite{verellocal}. There exists a directed transition $e_{ij}$ from $LO_i$ to $LO_j$ if it exists a solution $s$ such that $d(s, LO_i) \leq D$ and $h(s) = LO_j$, where the distance $d(s_i,s_j)$ can be measured in ``number of moves'' (i.e Hamming distance in the bit-flip operator case). The distance-threshold $D \in \mathbb{N}$ can be chosen accordingly to the applied perturbation; in this work, it is set to $D=2$. The weight $w_{ij}$ of such a transition is then: $w_{ij}= \sharp \{ s \in S ~|~ d(s, LO_i) \leq D \mbox{ and } h(s) = LO_j \}$, i.e. the number of paths at distance $D$ starting at $LO_i$ and reaching the basin of $LO_j$. This can be normalized by the number of solutions within reach w.r.t. the given distance threshold, i.e. $\sharp \{ s \in S ~|~ d(s, LO_i) \leq D\}$.

The weighted and directed graph $G=(V,E)$ having the set of vertices $V=\{LO_1, \dots, LO_{nv}\}$ and the set of edges $E=\{e_{ij} | w_{ij} > 0\}$, is the \emph{Local Optima Network} (LON) \cite{verellocal}.

\subsection{Iterated Local Search}
\label{sec:ILS}

Iterated local search is a relatively simple but successful algorithm. It operates by iteratively alternating between applying a move operator to  the incumbent solution and restarting local search from the perturbed solution. This search principle has been rediscovered multiple times, within different research communities and with different names \cite{Baxter1981,Martin1992}. The term {\em iterated local search} (ILS) was proposed in \cite{lourenco:2002}.  Algorithm~\ref{alg:ILS} outlines the procedure.

\begin{algorithm}[!h]
\caption{Iterated Local Search} \label{alg:ILS}
$s_0 \leftarrow \text{GenerateInitialSolution}$\;
$s^* \leftarrow \text{LocalSearch}(s_0)$\;
\Repeat{termination condition met}{
	$s' \leftarrow \text{Perturbation}(s^*)$\;
	$s'^{*} \leftarrow \text{LocalSearch}(s^{'})$\;
	$s^* \leftarrow \text{AcceptanceCriterion}(s^*,s'^{*})$\;
}
\end{algorithm}

In the present study, the base LocalSearch heuristic is the same best-improvement hill-climber of Algorithm~\ref{alg:HC}, which stops on a LO.
This heuristic  uses the single bit-flip move operator. Therefore,  a 2-bit-flip mutation is chosen as a Perturbation operator. When a different LO is found after that,
the search process accepts the move if its fitness is higher (we are assuming maximization). With these settings, ILS is  performing a first-improvement hill-climbing in the configuration space of the LON with the escape-edges at distance $D=2$, as defined in section~\ref{sec:LON}.

The search terminates at the global optimum, which for benchmark problems is known \textit{a priori}, or when reaching a pre-set limit of fitness evaluations $FE_{max}$.

\subsection{Performance Evaluation}
\label{sec:perfeval}

As the performance criterion, we selected the expected number of function evaluations to reach the global optimum (\emph{success}) after independent restarts of the ILS algorithm  (Algorithm~\ref{alg:ILS})~\cite{auger2005performance}. This measure accounts for both the success rate ($p_s \in (0,1]$) and the convergence speed. In theory, after $(N-1)$ unsuccessful runs stopped at $T_{us}$-steps and the final successful one running for $T_s$-steps, the total running time would be $T=\sum_{k=1}^{N-1}{(T_{us})}_{k}+T_s$. Taking the expectation and considering that $N$ follows a geometric distribution\footnote{probability distribution of the number $N-1$ of failures before the first success (\emph{Bernoulli trials})} with parameter $p_s$, it gives:
$$\mathbb{E}(T)=\left(\frac{1-p_s}{p_s}\right)\mathbb{E}(T_{us})+\mathbb{E}(T_{s})$$ where in the present case $\mathbb{E}(T_{us})=FE_{max}$, the ratio of successful to total runs is an estimator for $p_s$, and $\mathbb{E}(T_{s})$ can be estimated by the average running time of those successful runs.

The ILS variant detailed in Sec.~\ref{sec:ILS} is \emph{essentially incomplete}, i.e. there are soluble problem instances for which the success probability is $<1$ even in the limit of an infinite running time~\cite{hoos2005stochastic}.  Given the chosen acceptance criterion, the search will eventually get stuck. Indeed, out of the test runs, $1$ instance with $K=16$ and $3$ instances with $K=17$ were not  solved. This theoretical limitation could be overcome by performing as many random restart as to cover the whole search space, but such a solution is of limited practical interest for large problems. In the present study, the success performance has been estimated on all the instances that had been solved at least once.

\begin{table*}[!htb]
\begin{center}
 \caption{Group averages of all the observed variables, aggregated by the epistasis $K$ of the corresponding $NK$-landscape. Standard deviations are given in subscripts. $nv$ = number of vertices (Local Optima), $lo$ = average shortest path to reach the global optimum, $lv$ = average path length ($d_{ij}=1/w_{ij}$), $fnn$ = Spearman coefficient for the nearest-neighbors fitness-fitness correlation, $wii$ = average non-normalized weight of self-loops, $cc$ global clustering coefficient, $zout$ = average out-going degree, $y2$ = average weight disparity for out-going edges, $knn$ = degree assortativity, $ets$ = estimated time to succeed.\label{tab:stat}}
\resizebox{\textwidth}{!}{\begin{tabular}{lcccccccccc}
\toprule
\multicolumn{1}{l}{aggregate}&\multicolumn{1}{c}{nv}&\multicolumn{1}{c}{lo}&\multicolumn{1}{c}{lv}&\multicolumn{1}{c}{fnn}&\multicolumn{1}{c}{wii}&\multicolumn{1}{c}{cc}&\multicolumn{1}{c}{zout}&\multicolumn{1}{c}{y2}&\multicolumn{1}{c}{knn}&\multicolumn{1}{c}{ets ($\times 10^{4}$)}\tabularnewline
\midrule
$K=2$&$  43_{ 28}$&$33.5_{ 14}$&$ 187_{ 51}$&$0.703_{0.19}$&$ 105_{ 11}$&$0.425_{0.086}$&$ 6.9_{1.8}$&$0.392_{0.075}$&$0.155_{0.4}$&$2.16_{3.3}$\tabularnewline
$K=4$&$ 221_{ 39}$&$53.7_{ 12}$&$ 214_{ 15}$&$0.587_{0.07}$&$83.9_{  3}$&$0.263_{0.013}$&$14.3_{  1}$&$0.219_{0.016}$&$-0.536_{0.13}$&$8.39_{7.74}$\tabularnewline
$K=6$&$ 748_{ 70}$&$66.7_{ 13}$&$ 188_{4.8}$&$0.535_{0.041}$&$67.5_{1.9}$&$0.19_{0.005}$&$24.8_{0.86}$&$0.124_{0.0059}$&$-0.778_{0.035}$&$32.6_{36.1}$\tabularnewline
$K=8$&$1669_{ 73}$&$76.6_{9.1}$&$ 171_{1.9}$&$0.431_{0.025}$&$53.3_{0.88}$&$0.159_{0.0012}$&$35.7_{0.57}$&$0.0769_{0.002}$&$-0.856_{0.022}$&$51.8_{61.1}$\tabularnewline
$K=10$&$3148_{110}$&$90.7_{8.4}$&$ 166_{1.2}$&$0.342_{0.016}$&$40.7_{0.78}$&$0.143_{0.00085}$&$47.2_{0.57}$&$0.0491_{0.0011}$&$-0.904_{0.011}$&$81.5_{70.6}$\tabularnewline
$K=12$&$5270_{104}$&$ 108_{ 12}$&$ 170_{0.64}$&$0.255_{0.015}$&$30.8_{0.35}$&$0.133_{0.00054}$&$57.8_{0.39}$&$0.0334_{0.00046}$&$-0.928_{0.0093}$&$276_{544}$\tabularnewline
$K=14$&$8100_{121}$&$ 125_{8.6}$&$ 181_{0.6}$&$0.19_{0.011}$&$23.5_{0.25}$&$0.128_{0.00032}$&$66.9_{0.33}$&$0.0245_{0.00022}$&$-0.944_{0.0063}$&$300_{288}$\tabularnewline
$K=16$&$11688_{101}$&$ 146_{ 11}$&$ 197_{0.42}$&$0.143_{0.0073}$&$18.2_{0.11}$&$0.125_{0.00023}$&$74.6_{0.17}$&$0.0196_{7.8e-05}$&$-0.948_{0.0055}$&$414_{632}$\tabularnewline
$K=17$&$13801_{ 74}$&$ 156_{ 12}$&$ 205_{0.42}$&$0.133_{0.01}$&$16.1_{0.06}$&$0.125_{0.00021}$&$78.2_{0.13}$&$0.0179_{6.5e-05}$&$-0.944_{0.0063}$&$793_{844}$\tabularnewline
\bottomrule
\end{tabular}}
\end{center}
\end{table*}

The benchmark set consists of $NK$-landscapes with $N = 18$ and $K \in \{2,4,6,8,10,12,14,16,17\}$. Those are the largest possible parameter combinations for which we could afford the exhaustive extraction of the local optima networks. In order to minimize the influence of the random generation of landscapes, $30$ independent problem instances are considered for each combination of  $N$ and $K$, which accounts for a total of $270$ instances in the problem set. The function-evaluations limit is set to $1/5$ of $\sharp S$, i.e. $FE_{max} \simeq 5.2 \cdot 10^{4}$, success rate $p_s$ and running time of successful runs $T_s$ are estimated on $500$ random restarts per instance.

\section{Results}

\subsection{Descriptive Statistics}

Table~\ref{tab:stat} summarises the LON metrics. Results are grouped according to the $K$ value of the corresponding $NK$-landscapes, and present averages and standard deviations over $30$ independent realizations per group. The number of local optima $nv$, is a metric familiar to any description of a rugged landscape. The other metrics are particular to the complex-network perspective provided by the LON model~\cite{newman2003structure}.

From left to right in the table, $lo$ represents the average length of the shortest paths that reach the global optimum starting from any other local optimum. The cost associated to an edge  $\vec{ij}$ of the LON graph, is  $d_{ij}=1/w_{ij}$. This measure can be interpreted as the expected number of random perturbations for escaping $LO_i$ and entering exactly the basin of $LO_j$.  $lv$ gives the average path length for the whole graph, which accounts for the weighted network \emph{characteristic length}. $lo$ is intuitively  more directly related to the search difficulty. Indeed,  $lo$ increases steadily with the landscape ruggedness  $K$, whilst the trend with  $lv$ is less clear. As a possible explanation, the network growth in terms of nodes might be counteracted by a growth in nodes connectivity: the number of weighted outgoing transitions from a given $LO$, i.e. its out-degree in the LON, increases with $K$ (cf. column $zout$ in Tab.~\ref{tab:stat}).

Column $fnn$ measures the correlation between the fitness of a node and the weighted average of the fitness of its nearest neighbors~\cite{barth2005characterization}.  This is relevant as the ILS acceptance criterion takes the fitness values of LO into account. With respect to this metric, $NK$-landscapes behave at the LO level as they do at the solution level: they become the more and more uncorrelated with $K$ approaching $N-1$ (see column $fnn$). In general, it is expected that a high and positive $fnn$ correlation would help the search process.

Column $wii$ reports the average number of perturbations that remain in the same basin of attraction, which is a proxy for the basin size. The larger the basins, the more difficult it is to escape them. Table ~\ref{tab:stat} shows that  average $wii$  value decreases with the landscape ruggedness.

Column $cc$,  reports the \emph{clustering coefficient}~\cite{newman2003structure}, which  measures the ratio of connected triples in the LON graph. In a social network, this coefficient measures  the probability for one's friends to be friends among each other. In the LO networks, it provides an index of topological locality for the transitions between local optima. Table ~\ref{tab:stat} suggest that $cc$ values decrease steadily with increasing $K$, but so does the LON density~\cite{verellocal} and that might be the reason.

 Column $zout$ counts the number of transitions departing from a given LO. It is relevant to know whether all the transitions have the same  rate, or if there is a preferred direction. To this aim, the \emph{disparity} score $y2$ gauges the weight heterogeneity of out-going edges~\cite{barth2005characterization}. When all connections $e_{ij}$ leaving a given $LO_i$ have the same probability $w_{ij}$, the disparity  will be close to the inverse of the out-degree $1/zout_i$ otherwise it will be higher than this value. Column $y2$ shows that the disparity monotonically decreases with increasing $K$. From the point of view of a metaheuristic dynamic, a low disparity means that transitions are almost equiprobable. The LON topology does not preferentially guide the search trajectory, which makes the search process harder.

Column $knn$, reports  the nearest-neighbors degree correlation or \emph{assortativity}, a classical  description of the mixing pattern of nodes in a network. The assortativity measures the affinity to connect with high or low-degree neighbors. The LONs of $NK$-landscapes are strongly disassortative, i.e. LO with few connections tend to link to others with many, and conversely. The implications of this observation on the search difficulty are worth further investigation.

\begin{figure*}[!hp]
\begin{center}
\includegraphics[width=0.9\textwidth]{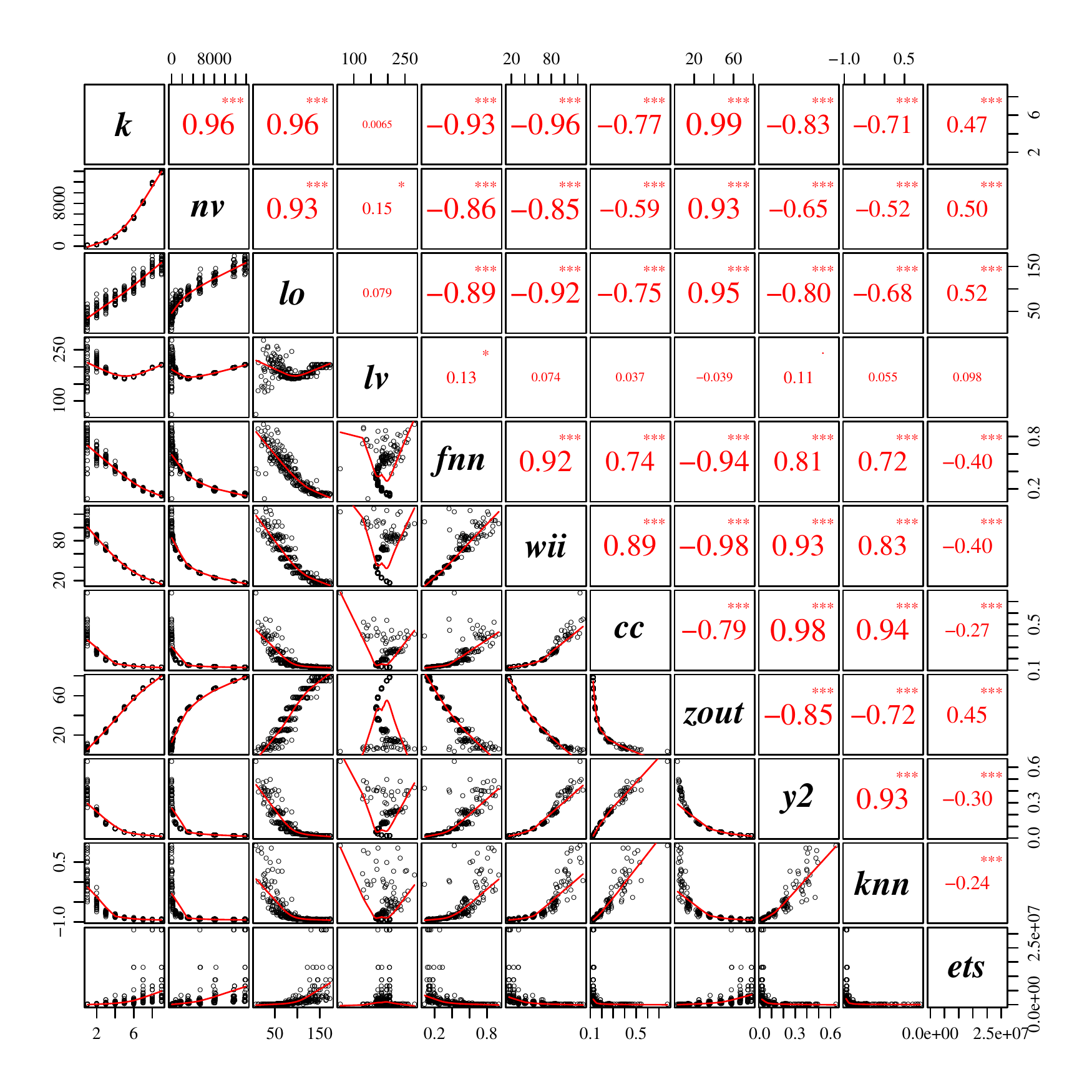}
\vspace{0.2cm}
\caption{Correlation matrix for all pairs of the observed variables, to be read from the diagonal. The lower panel displays scatter plots and smoothing splines for every possible pairing; the upper panel gives the corresponding Pearson correlation coefficient, with text size proportional to its absolute value. In the lower panel, the smoothing is done through locally-weighted polynomial regressions~\cite{cleveland1981lowess}. In the upper panel, the correlation coefficient is tested against the null hypothesis and the resulting p-value is symbolically encoded at the levels of $0.1$ ('), $0.05$ (*), $0.01$ (**), and $0.001$ (***).}
\label{fig:corgram}
\end{center}
\end{figure*}

\begin{figure*}[!hp]
\begin{center}
\includegraphics[width=0.9\textwidth]{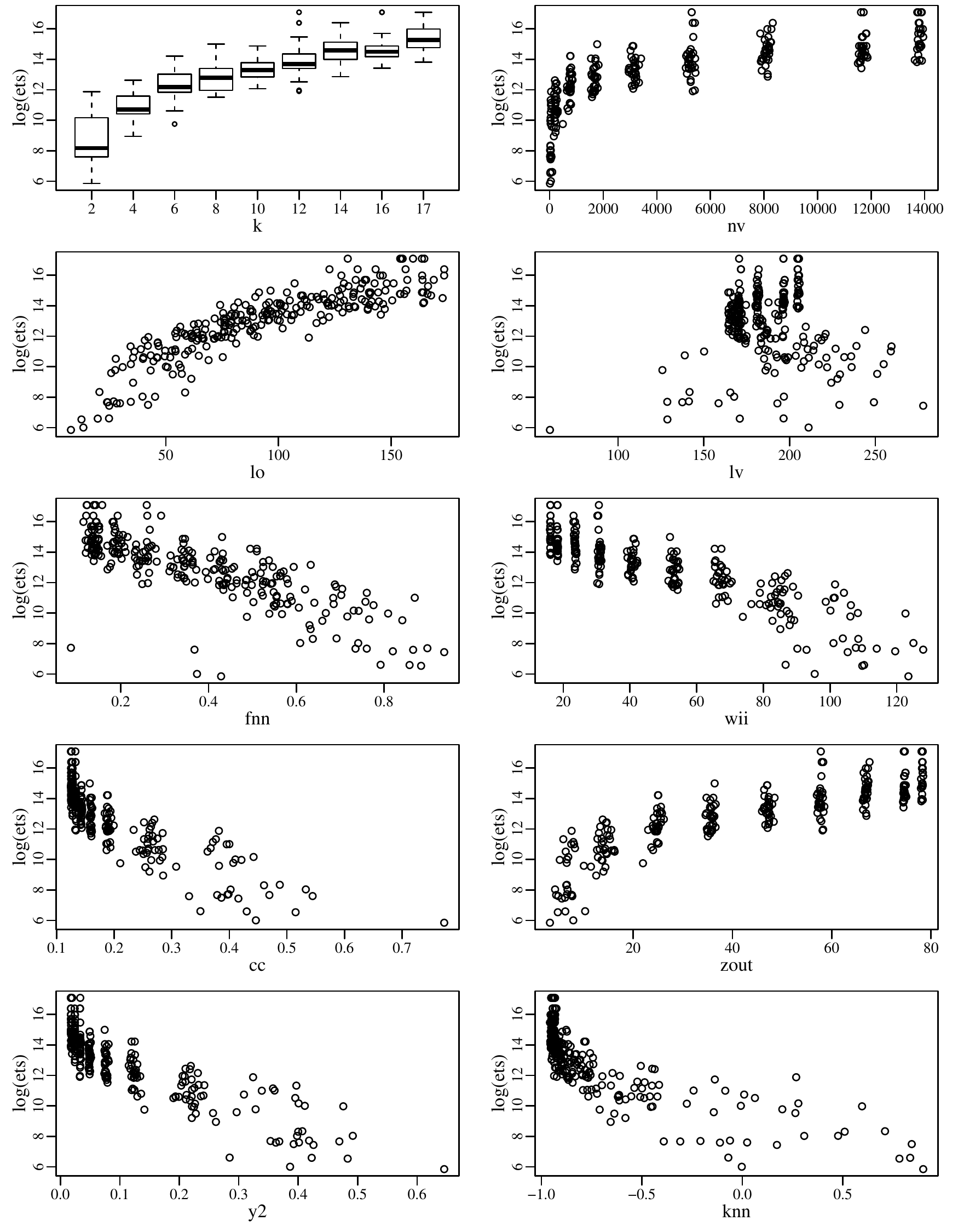}
\caption{Scatter plots of the logarithm of the estimated time to succeed against all measured variables (see labels on x-axis). The epistasis value $K$ is treated as a category, thus results are in a box-and-whisker diagram per group (fist plot on the top-left corner).}
\label{fig:splots}
\end{center}
\end{figure*}

The last column in Table~\ref{tab:stat} gives the values of the success performance indicator $\mathbb{E}(T)$ detailed in Sec~\ref{sec:perfeval}, which we abbreviate to $ets$ in the following. Clearly,  the expected running time increases with the landscape ruggedness (problem non-linearity) $K$. In order to analyze the  inter-correlations between all the studied metrics, Figure~\ref{fig:corgram} displays a \emph{correlogram} of the whole data set. The figure depicts  any possible pairing among the observed variables with a scatter plot in the panel below the diagonal, and the corresponding correlation coefficient  in the upper panel.  Table~\ref{tab:stat} corresponds to the first column  of the correlation matrix, i.e. the one showing how the different network metrics shape against the epistasis $K$.

 We are mostly interested in the correlations with the performance metric $ets$. Thus, the most relevant scatter plots are those in the last row of Figure~\ref{fig:corgram} together with the respective Pearson coefficients in the last column. By inspecting these values, the highest positive correlation is with the average length of the shortest path climbing to the global optimum ($r(ets,lo)\simeq0.52$). and with the total number of LO ($r(ets,nv)\simeq0.50$). This observation seems reasonable:
the more rugged the landscape  (i.e. the larger its LON), the higher the number of hops through the LON to reach the global optimum, and thus the longer the expected running time of a restarting local search. Conversely, the larger the LO basins, and the higher the nearest-neighbors fitness-fitness correlation, the shorter the running time (($r(ets,wii)\simeq r(ets,fnn) \simeq -0.40$)). The catch here,  is that those metrics are in turn correlated among themselves (e.g. $r(fnn,wii)\simeq0.92$), which prevents one from drawing relationships of causality from simple pair-wise correlations.

The last column plots also suggest a strong non-linearity of the performance estimator w.r.t. the LON metrics. In order to further analyze these relationships, Figure~\ref{fig:splots} zooms on the relevant scatter plots, and displays the logarithm of $ets$ as a function of the considered landscape measures.
The log-transformation allows to approach linearity, highlighting and confirming the results in the last column of \Cref{fig:corgram}. Namely, the relationships between $ets$ and all the metrics but $lv$ appear clearly. The picture  suggests a positive exponential trend with $lo$ and $zout$, and a negative exponential trend  with $fnn$, $wii$, $cc$, and $y2$. In the case of $nv$, the relation could also be close to power-law.

Since data are far from being normal bivariate, a more robust measure of association would be the rank-based Spearman's $\rho$ statistic, reported in Table~\ref{tab:rho}. These results complement the visual inspection of scatter plots and confirm the previous observations.

\begin{table}[!htbp]
\caption{Spearman's $\rho$ statistic for the correlation between $ets$ and the LON metrics (p-value $< 2.2e-16$ for all pairings).\label{tab:rho}}
\vspace{0.1cm}
\resizebox{0.48\textwidth}{!}{\begin{tabular}{ccccccccc}
\toprule
 nv     &lo     &lv    &fnn    &wii     &cc   &zout     &y2    &knn\tabularnewline
\midrule
$0.885$  &$0.915$  &$0.006$ &$-0.830$ &$-0.883$ &$-0.875$ &$0.885$ &$-0.883$ &$-0.850$\tabularnewline
\bottomrule
\end{tabular}}
\end{table}

\subsection{Statistical Modeling}

\Cref{fig:corgram,fig:splots}, along with \Cref{tab:rho}, address already some of the research questions asked in Section~\ref{sec:intro}, but do not provide an explanatory model for the algorithm performance as a function of the landscape features. To this end, we perform a multiple linear regression on the data, which has the general form:
\begin{equation}
y_i=\beta_0+\beta_1x_{i,1}+\beta_2x_{i,2}+\dots+\beta_px_{i,p}+\epsilon_i
\label{eq:lm}
\end{equation}
where the response variable $y$ in our case would be $ets$ and $p$ different predictors $x_{j}$ are to be chosen among the LON metrics; $\epsilon$ is the usual random noise term.

The least square regression produces estimates $\hat{\beta_j}$ for the $\beta_j$ model coefficients; the difference between the predicted values and the actual, observed values, are the regression residuals:
$e_i=y_i-\hat{y_i}=y_i-\hat{\beta_1}x_{i1}-\hat{\beta_2}x_{i2}-\dots-\hat{\beta_p}x_{ip}$.

The difficulty of this analysis is  that there are several possible explanatory variables, which are in turn intercorrelated. In consequence,  some of them could have
a confounding effect on the regression. In general, when confounders are known, measurable, and measured, it is a good practice to include them in the model. We, therefore,  start by fitting the following formula:
\begin{equation}
\log(ets)=\beta_0+ \beta_1 k + \beta_2 \log(nv)+\beta_2lo+\dots+\beta_{10} knn+\epsilon
\label{eq:lm1}
\end{equation}
where, w.r.t the general expression~\ref{eq:lm}, the response and one of the predictors have been log-trans\-formed in order to better approach linearity, as seen in \Cref{fig:splots}.
Moreover, the variable $k$ is qualitative and enters the model as a fixed effect, which translates in one appropriate dummy variable for each class $K=2,K=4,\dots, K=17$.

The summary statistics for this model are reported in Table~\ref{tab:lm1}. In the table caption, the multiple $R^2$ represents the proportion of variance explained by the linear regression.  $R^2$  would be equal to $1$ if all observed data points were lying on the regression plane. When comparing models with a different number of predictors, the adjusted $R^2$ should  be used instead.
The $F$ statistic is the ratio of the variance explained by the parameters in the model, to the residual or unexplained variance. The $p$-value is the probability of achieving an $F$ that large under the null hypothesis of no effect~\cite{chambers1992statistical}. 

The estimated coefficient $\hat{\beta_j}$ and their estimated standard error $\hat{\sigma_j}$ are given in the $2^{nd}$ and $3^{rd}$ columns, respectively.  Their ratio is the $t$-statistic ($4^{th}$ column) that is used to calculate a $p$-value for the significance of the estimation (last column).

\begin{table}[!htbp]
\begin{center}
\caption{Summary statistics of the linear regression model on all variables. Residual standard error: $0.8702$ on $248$ degrees of freedom
  ($4$ observations deleted due to missingness).
Multiple R-squared: $0.8585$,	Adjusted R-squared: $0.8488$.
F-statistic: $88.52$ on $17$ and $248$ DF,  $p$-value: $< 2.2e-16$.\label{tab:lm1}}
\resizebox{0.48\textwidth}{!}{\begin{tabular}{lrrrr}
\toprule
\multicolumn{1}{l}{summary}&\multicolumn{1}{c}{Estimate}&\multicolumn{1}{c}{Std. Error}&\multicolumn{1}{c}{$t$ value}&\multicolumn{1}{c}{Pr(\textgreater $|t|$)}\tabularnewline
\midrule
(Intercept)&$ 16.06966$&$7.39640$&$ 2.1726$&$3.08e-02$\tabularnewline
k4&$  0.01542$&$0.63987$&$ 0.0241$&$9.81e-01$\tabularnewline
k6&$ -1.08926$&$1.39976$&$-0.7782$&$4.37e-01$\tabularnewline
k8&$ -3.14529$&$2.46616$&$-1.2754$&$2.03e-01$\tabularnewline
k10&$ -5.67316$&$3.76828$&$-1.5055$&$1.33e-01$\tabularnewline
k12&$ -8.19327$&$5.04638$&$-1.6236$&$1.06e-01$\tabularnewline
k14&$-10.34765$&$6.17715$&$-1.6751$&$9.52e-02$\tabularnewline
k16&$-12.85523$&$7.14074$&$-1.8003$&$7.30e-02$\tabularnewline
k17&$-13.40456$&$7.59325$&$-1.7653$&$7.87e-02$\tabularnewline
log(nv)&$ -1.91370$&$1.12656$&$-1.6987$&$9.06e-02$\tabularnewline
lo&$  0.04882$&$0.00499$&$ 9.7919$&$2.37e-19$\tabularnewline
lv&$  0.00198$&$0.00376$&$ 0.5265$&$5.99e-01$\tabularnewline
fnn&$  0.54148$&$0.89574$&$ 0.6045$&$5.46e-01$\tabularnewline
wii&$ -0.00302$&$0.02739$&$-0.1104$&$9.12e-01$\tabularnewline
cc&$ -7.22853$&$5.00042$&$-1.4456$&$1.50e-01$\tabularnewline
zout&$  0.29514$&$0.15838$&$ 1.8636$&$6.36e-02$\tabularnewline
y2&$ -3.46837$&$5.00914$&$-0.6924$&$4.89e-01$\tabularnewline
knn&$ -0.88961$&$0.49062$&$-1.8132$&$7.10e-02$\tabularnewline
\bottomrule
\end{tabular}}
\end{center}
\end{table}

In this initial model, the average length of paths to the global optimum $lo$ is the only predictor with a regression coefficient
that is statistically-significant at the $0.05$ threshold ($\beta_{lo}=0.04882$, $p$-value = $2.37e-19$).

Therefore, we proceed to perform a step-wise model selection by \emph{backward elimination}~\cite{mass}. From the initial formula, at
each step we compute what change in the fit could be produced by dropping each predictor in turn, and then we eliminate the one that
minimizes the AIC score of the resulting model~\cite{sakamoto1987akaike}. By iterating until all predictors become significant,
we obtain the final model:
\begin{equation}
\log(ets)=\beta_0+ \beta_{lo} lo + \beta_{zout} zout + \beta_{y2} y2 + \beta_{knn} knn+\epsilon
\label{eq:lm2}
\end{equation}
which is detailed in Table~\ref{tab:lm2}.

\begin{table}[!htbp]
\begin{center}
\caption{Summary statistics of the final linear regression model. Residual standard error: $0.8751$ on $261$ degrees of freedom
  ($4$ observations deleted due to missingness).
Multiple R-squared: $0.8494$,	Adjusted R-squared: $0.8471$.
F-statistic:   $368.1$ on $4$ and $261$ DF,  $p$-value: $< 2.2e-16$.\label{tab:lm2}}
\vspace{0.2cm}
\begin{tabular}{lrrrr}
\toprule
\multicolumn{1}{l}{summary}&\multicolumn{1}{c}{Estimate}&\multicolumn{1}{c}{Std. Error}&\multicolumn{1}{c}{$t$ value}&\multicolumn{1}{c}{Pr(\textgreater $|t|$)}\tabularnewline
\midrule
(Intercept)&$10.3838$&$0.58512$&$17.75$&$9.24e-47$\tabularnewline
lo&$ 0.0439$&$0.00434$&$10.11$&$1.67e-20$\tabularnewline
zout&$-0.0306$&$0.00831$&$-3.68$&$2.81e-04$\tabularnewline
y2&$-7.2831$&$1.63038$&$-4.47$&$1.18e-05$\tabularnewline
knn&$-0.7457$&$0.40501$&$-1.84$&$6.67e-02$\tabularnewline
\bottomrule
\end{tabular}
\end{center}
\vspace{-0.2cm}			
\end{table}

This final model is able to explain $84.94\%$ of the variance observed in $\log(ets)$ with a linear regression on four variables
that are all LON network metrics. Among these metrics, the length of the paths to the global optimum, and the weight disparity have
the highest relative importance~\cite{relimpo}.

\begin{figure}[!hb]
\begin{center}
\vspace{0.2cm}
\includegraphics[width=0.4\textwidth]{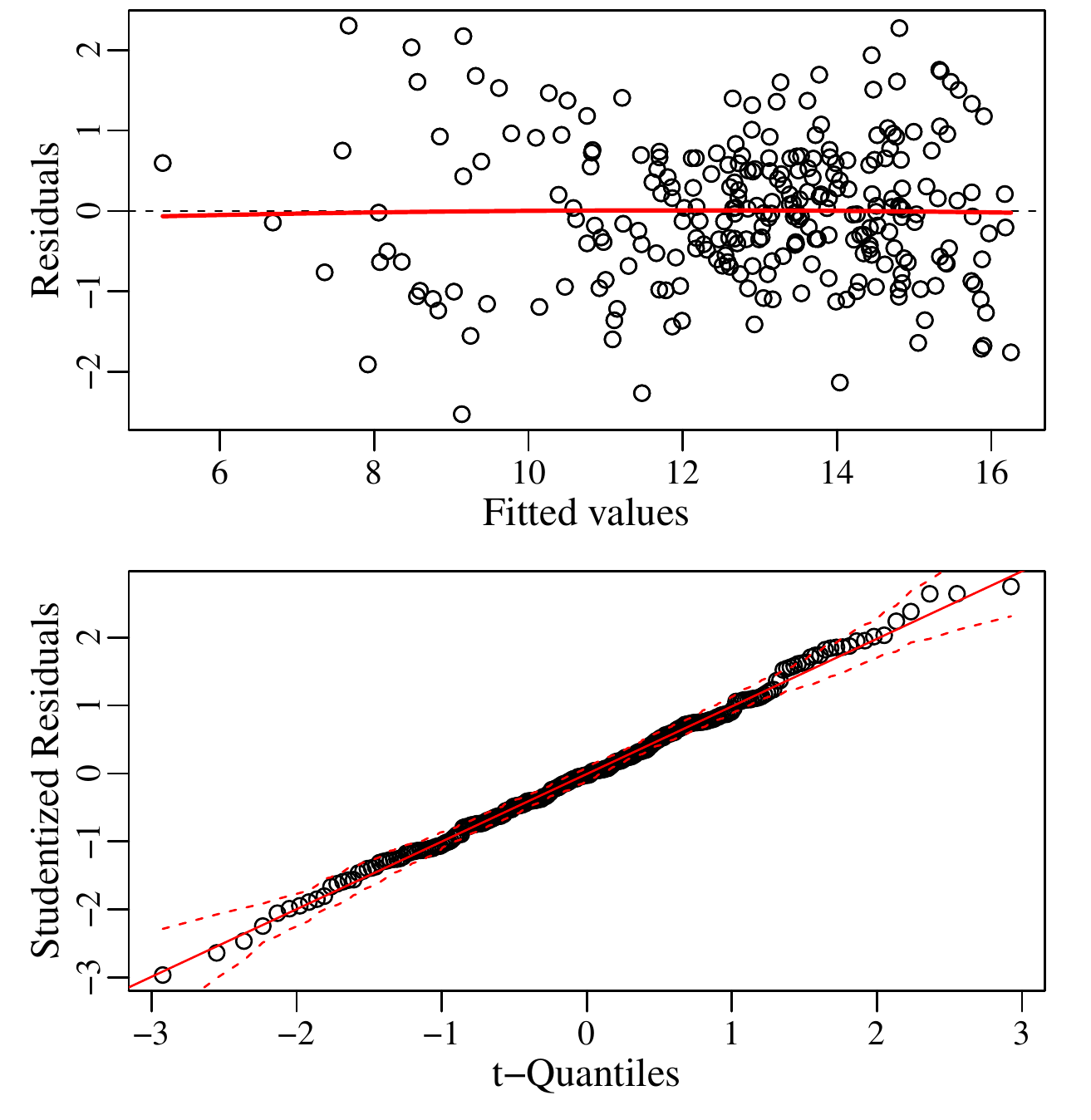}
\caption{Top: residual plot, to asses the hypothesis of zero-mean and constant-variance of the regression residuals (circle dots) around the fitted values (dotted line); no visually-significant deviation appears (red smooth line). Bottom: quantile-quantile comparison of the studentized regression residuals (circle dots) against the theoretical quantiles (red thick line), to inspect the distribution of residuals; no significant deviation from normality appears (confidence intervals as dotted red lines).}
\label{fig:resplot}
\end{center}
\end{figure}

\begin{figure}[!htb]
\begin{center}
\vspace{0.3cm}
\includegraphics[width=0.41\textwidth]{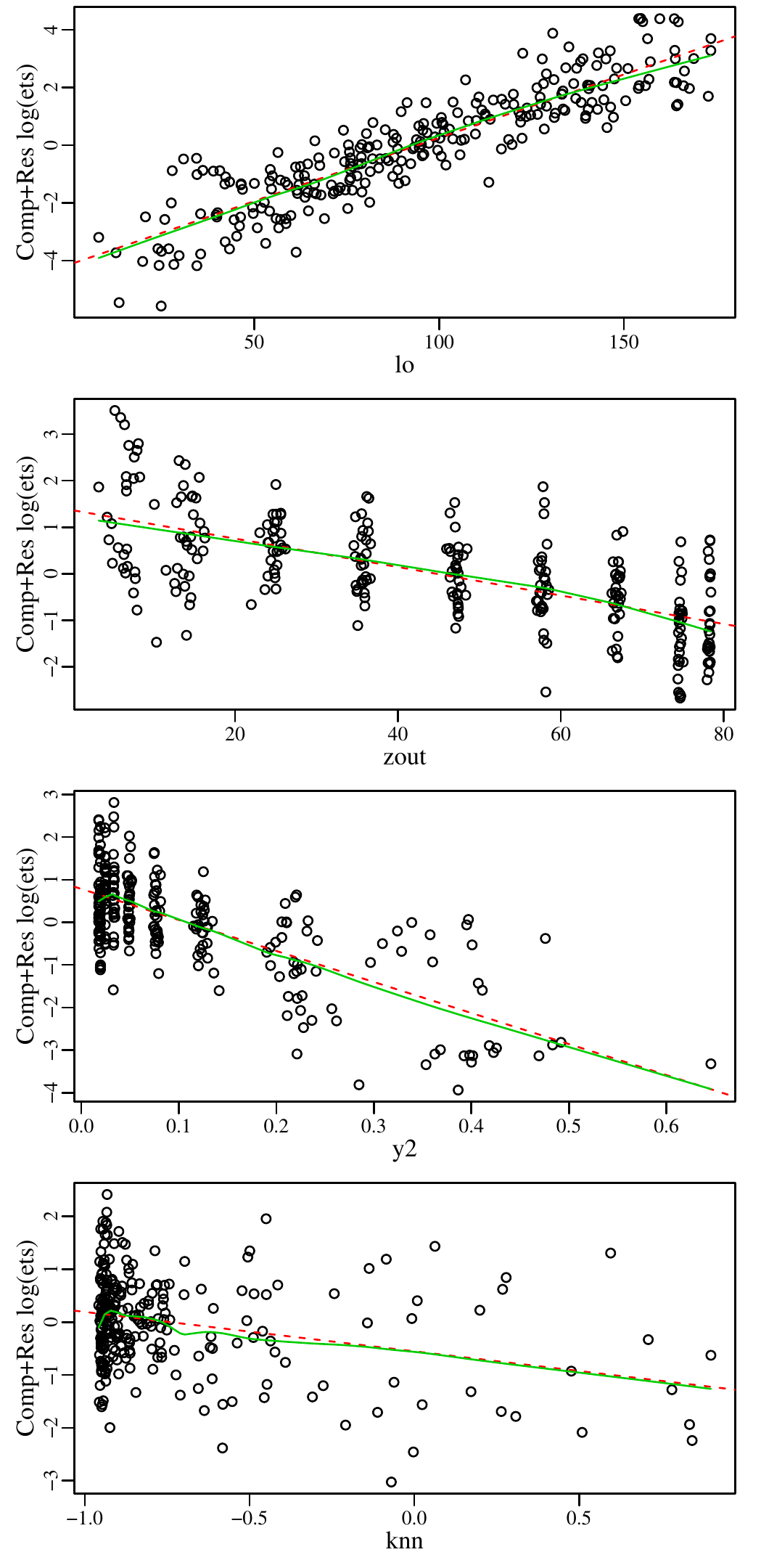}
\caption{Component+residual plots for the linear regression model. Circle dots are, for each observation, the corresponding residual error from the regression plus the value fitted by one explanatory variable alone, plotted against that variable (see labels on x-axis). No significant deviation from linearity appears (smooth green line against the red dotted line of the partial regression).}
\label{fig:compresplot}
\end{center}
\end{figure}

Without a check on the model assumptions, this would remain an observational study and could not be used to make predictions.
To this end, a combination of parametric tests (not reported for space reasons) provided a positive confirmation~\cite{gvlma}. However,  a visual diagnostics is more informative. In particular, Figure~\ref{fig:resplot} helps to assess if the regression residuals follow a normal distribution with zero mean and
homogeneous variance, whereas Figure~\ref{fig:compresplot} displays the contributions to the model of each predictor in turn,
highlighting possible violations of the linearity hypothesis~\cite{car}.
All assumptions seem  acceptable. Therefore, formula~\ref{eq:lm2} could be used to make inferences. In other words,  formula~\ref{eq:lm2} coefficients
can be interpreted as conditional expectations for the average change in the response when one predictor undergoes a unitary
change, and all the others remain fix. Since the dependent variable is log-transformed, this effect would be multiplicative.

The only important limitation of the proposed model is~\emph{multicollinearity}: due to the complex intercorrelations among LON metrics (cf. \Cref{fig:corgram}),
predictors are not really independent. This does not invalidate the multiple linear regression analysis, but it inflates the variance of its coefficients
and  makes it harder to disentangle their respective contributions~\cite{car}.

\balance

\section{Conclusions}

This article  explored correlations between local optima network features and the performance of a stochastic local search algorithm running on the underlying combinatorial optimization problem.  The  NK family of landscapes and the iterated local search meta-heuristic were considered.  It has been shown in previous work e.g.~\cite{pre08,verel2010local} that some features of the LON networks  are related to the $NK$-landscapes ruggedness, and thus to problem difficulty. However, no statistically testable model was presented.  The contribution of this study was to investigate, with a statistically-sound approach, which  features of the LONs have a strong influence on the search performance, expressed as expected running times to success. The results obtained through the use of a  multiple linear regression
model show that some LON metrics are more important than others. These are: the average length of the shortest paths to the optimum, the average out-degree, the average disparity, and the degree assortativity. This study confirms and provides significant evidence that LON modeling is a compressed-but-relevant view of the fitness landscape, and can be used to understand and predict search difficulty.

It is worth noticing that some network metrics can be estimated without knowing the global optimum beforehand, such as the average out-degree, the fitness-fitness correlation, the average disparity and assortativity. Using these metrics and an adequate statistical model, as we have done in this work, opens up exciting possibilities. With  standard sampling methods, larger  search spaces could be studied. Thereafter, using the performance model based on the estimated LON metrics, the search heuristic parameters or its operators can be off-line tuned, or even on-line controlled. We plan to continue working in this direction and  to extend this analysis to other combinatorial problems such as QAP.

\section{Acknowledgments}
Fabio Daolio and Marco Tomassini
gratefully acknowledge the Swiss National Science Foundation for financial support under grant number 200021-124578.

\bibliographystyle{abbrv}
\small

\begin{thebibliography}{}

\end{thebibliography}


\begin{thebibliography}{10}

\bibitem{auger2005performance}
A.~Auger and N.~Hansen.
\newblock Performance evaluation of an advanced local search evolutionary
  algorithm.
\newblock In {\em Evolutionary Computation, 2005. The 2005 IEEE Congress on},
  volume~2, pages 1777--1784. IEEE, 2005.

\bibitem{barth2005characterization}
M.~Barth{\'e}lemy, A.~Barrat, R.~Pastor-Satorras, and A.~Vespignani.
\newblock Characterization and modeling of weighted networks.
\newblock {\em Physica A: Statistical Mechanics and its Applications},
  346(1):34--43, 2005.

\bibitem{Baxter1981}
J.~Baxter.
\newblock Local optima avoidance in depot location.
\newblock {\em Journal of the Operational Research Society}, 32:815--819, 1981.

\bibitem{chambers1992statistical}
J.~Chambers, T.~Hastie, et~al.
\newblock Linear models.
\newblock In {\em Statistical models in S}, chapter~4. Chapman \& Hall London,
  1992.

\bibitem{cleveland1981lowess}
W.~Cleveland.
\newblock Lowess: A program for smoothing scatterplots by robust locally
  weighted regression.
\newblock {\em The American Statistician}, 35(1):54--54, 1981.

\bibitem{daolio2011communities}
F.~Daolio, M.~Tomassini, S.~V{\'e}rel, and G.~Ochoa.
\newblock Communities of minima in local optima networks of combinatorial
  spaces.
\newblock {\em Physica A Statistical Mechanics and its Applications},
  390:1684--1694, 2011.

\bibitem{daolio2010local}
F.~Daolio, S.~Verel, G.~Ochoa, and M.~Tomassini.
\newblock {L}ocal {O}ptima {N}etworks of the {Q}uadratic {A}ssignment
  {P}roblem.
\newblock In {\em Evolutionary Computation (CEC), 2010 IEEE Congress on}, pages
  1--8. IEEE, 2010.

\bibitem{car}
J.~Fox and S.~Weisberg.
\newblock {\em An {R} Companion to Applied Regression}.
\newblock Sage, Thousand Oaks {CA}, second edition, 2011.

\bibitem{glover1986future}
F.~Glover.
\newblock Future paths for integer programming and links to artificial
  intelligence.
\newblock {\em Computers \& Operations Research}, 13(5):533--549, 1986.

\bibitem{relimpo}
U.~Groemping.
\newblock Relative importance for linear regression in {R}: The package
  relaimpo.
\newblock {\em Journal of Statistical Software}, 17(1):1--27, 2006.

\bibitem{hoos2005stochastic}
H.~Hoos and T.~St{\"u}tzle.
\newblock {\em Stochastic local search: Foundations and applications}.
\newblock Morgan Kaufmann, 2005.

\bibitem{kauffman93}
S.~A. Kauffman.
\newblock {\em The Origins of Order}.
\newblock Oxford University Press, New York, 1993.

\bibitem{lourenco:2002}
H.~R. Louren{\c c}o, O.~Martin, and T.~St{\" u}tzle.
\newblock Iterated local search.
\newblock In {\em Handbook of Metaheuristics}, volume~57 of {\em International
  Series in Operations Research \& Management Science}, pages 321--353. Kluwer
  Academic Publishers, 2002.

\bibitem{Martin1992}
O.~Martin, S.~W. Otto, and E.~W. Felten.
\newblock Large-step {M}arkov chains for the {T}{S}{P} incorporating local
  search heuristics.
\newblock {\em Operations Research Letters}, 11(4):219--224, 1992.

\bibitem{newman2003structure}
M.~Newman.
\newblock The structure and function of complex networks.
\newblock {\em SIAM review}, pages 167--256, 2003.

\bibitem{gecco08}
G.~Ochoa, M.~Tomassini, S.~Verel, and C.~Darabos.
\newblock A study of {NK} landscapes' basins and local optima networks.
\newblock In {\em Genetic and Evolutionary Computation Conference, GECCO 2008},
  pages 555--562. ACM, 2008.

\bibitem{ppsn10}
G.~Ochoa, S.~Verel, and M.~Tomassini.
\newblock First-improvement vs. best-improvement local optima networks of nk
  landscapes.
\newblock In {\em Parallel Problem Solving from Nature - PPSN XI}, volume 6238
  of {\em Lecture Notes in Computer Science}, pages 104--113. Springer, 2010.

\bibitem{Pelikan:2010:NLP:1830483.1830606}
M.~Pelikan.
\newblock Nk landscapes, problem difficulty, and hybrid evolutionary
  algorithms.
\newblock In {\em Proceedings of the 12th annual conference on Genetic and
  evolutionary computation}, GECCO '10, pages 665--672, New York, NY, USA,
  2010. ACM.

\bibitem{gvlma}
E.~A. Pena and E.~H. Slate.
\newblock {\em gvlma: Global Validation of Linear Models Assumptions}, 2010.
\newblock R package version 1.0.0.1.

\bibitem{reidys2002combinatorial}
C.~Reidys and P.~Stadler.
\newblock {Combinatorial landscapes}.
\newblock {\em SIAM review}, 44(1):3--54, 2002.

\bibitem{sakamoto1987akaike}
Y.~Sakamoto and G.~Kitagawa.
\newblock {\em Akaike information criterion statistics}.
\newblock Kluwer Academic Publishers, 1987.

\bibitem{pre08}
M.~Tomassini, S.~Verel, and G.~Ochoa.
\newblock Complex-network analysis of combinatorial spaces: The {NK} landscape
  case.
\newblock {\em Phys. Rev. E}, 78(6):066114, 2008.

\bibitem{mass}
W.~N. Venables and B.~D. Ripley.
\newblock {\em Modern Applied Statistics with S}.
\newblock Springer, New York, fourth edition, 2002.
\newblock ISBN 0-387-95457-0.

\bibitem{verellocal}
S.~Verel, F.~Daolio, G.~Ochoa, and M.~Tomassini.
\newblock {Local Optima Networks with Escape Edges}.
\newblock In {\em {Procedings of International Conference on Artificial
  Evolution (EA-2011)}}, pages 10 -- 23, Angers, France, Oct 2011.

\bibitem{verel2010local}
S.~Verel, G.~Ochoa, and M.~Tomassini.
\newblock Local optima networks of {NK} landscapes with neutrality.
\newblock {\em Evolutionary Computation, IEEE Transactions on}, 6(15):783 --
  797, 2011.

\end{thebibliography}

\end{document}